%% file: main.tex
\documentclass[conference]{IEEEtran}
\IEEEoverridecommandlockouts
\usepackage{cite}
\usepackage{amsmath,amssymb,amsfonts}
\input{math_commands.tex}

\usepackage{algorithm}
\usepackage{multirow}
\usepackage{url}
\usepackage{float}
\usepackage{placeins} 
\usepackage{blindtext}
\usepackage[bookmarks=false]{hyperref}
\usepackage{algorithmic}
\usepackage{graphicx}
\usepackage{textcomp}
\usepackage{xcolor}
\def\BibTeX{{\rm B\kern-.05em{\sc i\kern-.025em b}\kern-.08em
    T\kern-.1667em\lower.7ex\hbox{E}\kern-.125emX}}
\begin{document}

\title{Interpretable Time-series Classification \\ on Few-shot Samples
}

\author{\IEEEauthorblockN{Wensi Tang}
\IEEEauthorblockA{\textit{Centre for AI, FEIT} \\
\textit{University of Technology Sydney}\\
Sydney, Australia \\
Wensi.Tang@student.uts.edu.au}
\and
\IEEEauthorblockN{Lu Liu}
\IEEEauthorblockA{\textit{Centre for AI, FEIT} \\
\textit{University of Technology Sydney}\\
Sydney, Australia \\
Lu.Liu-10@student.uts.edu.au}
\and
\IEEEauthorblockN{Guodong Long}
\IEEEauthorblockA{\textit{Centre for AI, FEIT} \\
\textit{University of Technology Sydney}\\
Sydney, Australia \\
Guodong.Long@uts.edu.au}
}

\maketitle

\begin{abstract}
Recent few-shot learning works focus on training a model with prior meta-knowledge to fast adapt to new tasks with unseen classes and samples. 
However, conventional time-series classification algorithms fail to tackle the few-shot scenario. 
Existing few-shot learning methods are proposed to tackle image or text data, and most of them are neural-based models that lack interpretability.
This paper proposes an interpretable neural-based framework, namely \textit{Dual Prototypical Shapelet Networks (DPSN)} for few-shot time-series classification, which not only trains a neural network-based model but also interprets the model from dual granularity: 1) global overview using representative time series samples, and 2) local highlights using discriminative shapelets.
In particular, the generated dual prototypical shapelets consist of representative samples that can mostly demonstrate the overall shapes of all samples in the class and discriminative partial-length shapelets that can be used to distinguish different classes. We have derived 18 few-shot TSC datasets from public benchmark datasets and evaluated the proposed method by comparing with baselines. The DPSN framework outperforms state-of-the-art time-series classification methods, especially when training with limited amounts of data. Several case studies have been given to demonstrate the interpret ability of our model.

\end{abstract}


\section{Introduction}
Training a classification model with very few labeled training samples, namely few-shot classification, has attracted broad interest recently. In particular, the few-shot classification model should be able to categorize unseen samples to a class according to the generalized concepts and knowledge that is extracted from very few seen examples (training samples) in this class.

Time-series classification has been broadly applied in intelligent-based real-world applications, including heart disease diagnoses (e.g., ECG200 data set), motion detection (e.g., GunPoint data set), traffic analysis (e.g., MelbournePedestrian data set), and earthquake prediction (e.g., Earthquakes data set), etc. However, conventional time-series classification methods fail to tackle the few-shot scenario. Solving the few-shot time-series classification problem will not only result in the development of a fast adaptive model for time-series data but also will benefit applications that have difficulty in acquiring enough labeled samples. For instance, both using functional MRI to provide a voxel-level damage assessment for an Alzheimer’s disease patient~\cite{busatto2008voxel} and using a mobile EEG headset to provide audio feedback on a learner’s attention level ~\cite{sun2017effects} suffer from insufficient supervision information.

\begin{figure}
\centering 
\includegraphics[clip,scale=0.4]{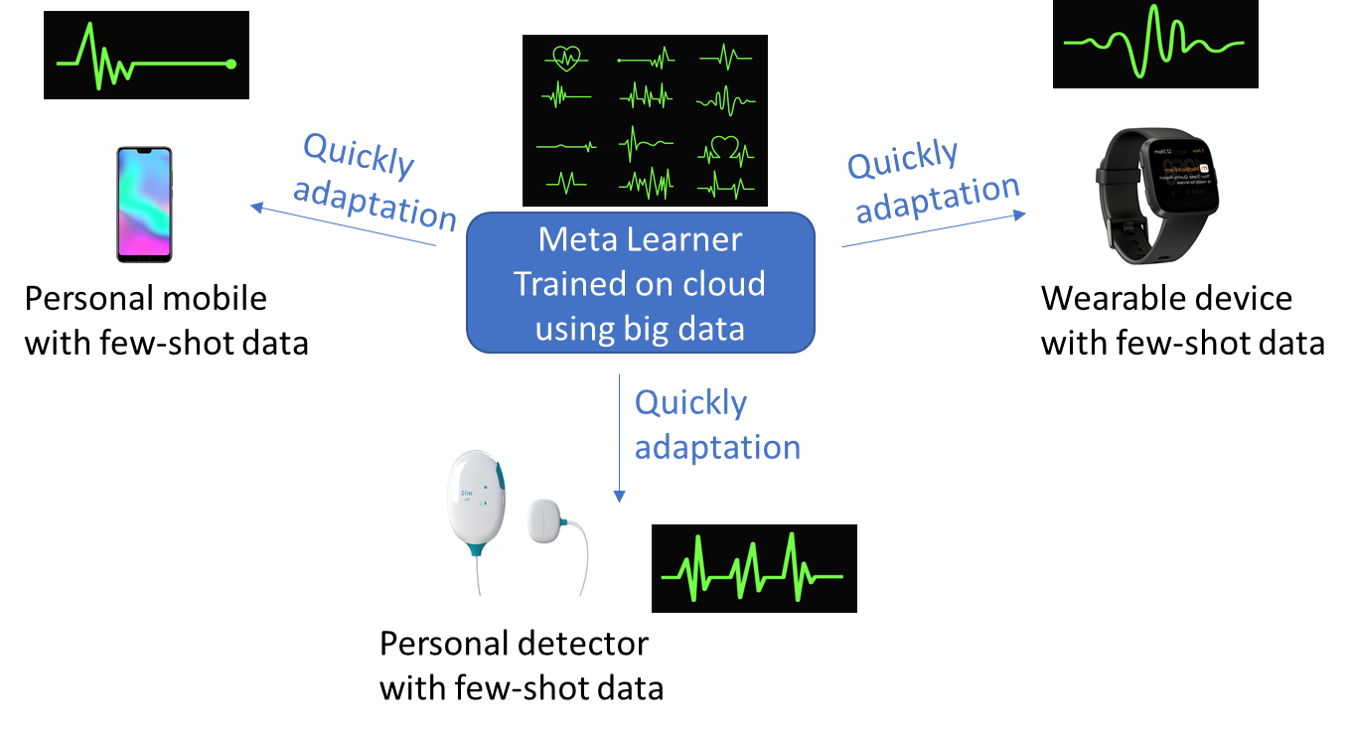}
\caption{Application scenarios for few-shot time-series classification on wearable devices and smartphones. Each end-user can use his or her own data to train a private intelligent model while preserving privacy.
}
\label{fig:motivation_for_few-shot}
\end{figure}

Recently proposed neural network-based few-shot learning models achieve state-of-the-art performance in many image and text classification related tasks~\cite{chen2018a,snell2017prototypical,NIPS2017_7244}. Few-shot time series classification is a new area with a promising future due to the following: firstly, with the development of the Internet of Things (IoT) and edge computing, numerous edge devices are deployed to continuously collect environment data which is usually in time series format, then these devices can make intelligent decisions according to its own data~\cite{long2012tcsst}. The smart applications in edge devices can be regarded as a typical problem of few-shot time-series classification~\cite{jiang2020decentralized}. Secondly, the rising privacy concerns on personal data, especially the European Union published General Data Protection Regulation (GDPR), will push the intelligent service from big data centering model towards the few-shot distributed personal model. As shown in Figure~\ref{fig:motivation_for_few-shot}, each end-user should have a private model that is trained with the user's own few-shot data. Lastly, the wide-spreading use of wearable devices and smartphones enables the development of an intelligent service which is based on the end-users' daily time-series data, e.g. medical time series, moving speed, and sleeping quality~\cite{wahbeh2007binaural,gao2014analysis}.

Model interpretability is another challenge when addressing this problem for the reason that most neural-based models are black-box models that lack interpretability. A reasonable interpretation is critically important to convince end-users that the trained few-shot model has a good generalisation capability even though the model is trained with very few training samples. 

\begin{figure*}[ht] 
\centering
\includegraphics[width=0.75\linewidth]{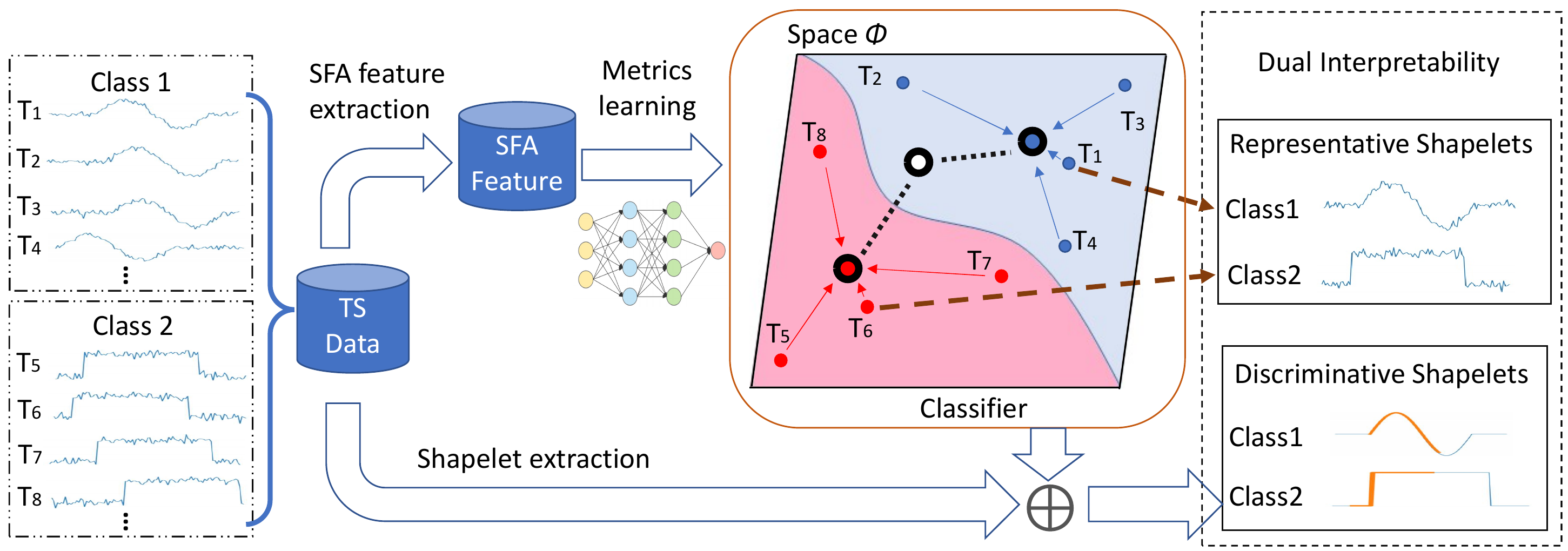}
\caption{The framework of DPSN. The framework contains three components: 1) feature extraction will transform the Time series into an SFA feature and Shapelet feature; 2) The classification part will build a prototype of each class by aggregating SFA feature via a metric learning neural network; 3) The dual interpretability has two-fold explanations including the representative shapelet which was identified by a prototype from the classification part, and discriminative shapelets which can be learned by combining information from shapelet feature and prototype.}
\label{fig:framework}
\end{figure*}

This paper proposes a novel Dual Prototypical Shapelet Network (DPSN) that can simultaneously solve the aforementioned few-shot time-series classification problem and address the model interpretability challenge. In particular, the classification model is built on an end-to-end framework that includes a KNN-based lazy learner, and a neural layer which is applied to learn the metrics to transform time series data into a new space. Dual interpretation of the classifier will be generated by two different model interpretability techniques. The first type of interpretation is a selected time-series example, namely a representative shapelet, to be used as a prototype to demonstrate the overall shape of all the samples in a class. The other interpretation output is a sub-sequence of a time-series example, namely discriminative shapelet, that can be used to distinguish different classes. The framework of DPSN is shown in Figure \ref{fig:framework}.

The paper’s main contributions are summarised as follows: 
\begin{itemize}
\item The first study solves the few-shot time-series classification task with neural-based methods.
\item It proposes a novel dual prototypical shapelet network that not only outperforms several state-of-the-art time-series classification methods in a few-shot scenario but can also generate interpretation.
\item  The proposed dual interpretation method is a novel interpretability one for time-series classification.
\end{itemize}
The remainder of the paper is organized as follows. Section~\ref{sec:related-works} introduces the related work. The methodology is described in Section~\ref{sec:method} followed by Section~\ref{sec:experiments} with the experiment results, and the final part is conclusion in Section~\ref{sec:Conclusions}.

\section{Related Works} \label{sec:related-works}
\subsection{Time Series Classification}
Time series classification is a classical research topic with various solutions. For example, the window method is widely used to extract distinctive sub-sequences to avoid the alignment issue. New distance metrics, such as DTW distance ~\cite{ratanamahatana2005three}, are introduced to mitigate the influences caused by the length variance of the discriminate sub-sequence and alignment problem. Discrete Fourier Transform (DFT) is used to minimize the alignment and noise problem by mapping time series data into a frequency domain.  Other sophisticated methods include a tree-based ensemble method (random forest) and SVM with a quadratic and cubic kernel, etc\cite{schafer2015boss}.
Based on the techniques mentioned above, many algorithms have been proposed. From the University of California, Riverside (UCR) time series classification archive  ~\cite{UCRArchive2018}, WEASEL, ST and BOSS are three SOTA traditional algorithms for TSC. There are also numerous deep learning-based methods to solve TSC tasks \cite{fawaz2019deep,tang2020rethinking}.

\subsection{Few-shot Learning}
In the pre-deep learning era, data is not collected systematically so that few-shot learning is proposed to target one task with limited samples. Bayesian model is used to encode the priors, which is a reflection of the object category distribution~\cite{fei2006one}. Attribute annotations of every category can help to alleviate the few-shot challenge even if no sample is available for every class, which is called zero-shot learning~\cite{lampert2013attribute}.

Data repositories are accumulating gradually. With the advent of datasets such as ImageNet~\cite{imagenet} and the improvement of deep neural networks such as ResNet~\cite{He_2016_CVPR}, comes the big data era.
Few-shot learning in the big data era focuses on the setting of continual learning or life-long learning, where new classes or new tasks with limited samples are considered as the main challenge for few-shot learning. To meet this challenge, researchers apply the idea of meta learning where a meta-learner extracts the common knowledge over various tasks or classes so that this meta knowledge can be transferred to new tasks with few samples.
The meta learner can grasp a shared compelling initialization for learner models~\cite{finn2017model}, a shared metric space for K-nearest neighbors search~\cite{snell2017prototypical}, a graph~\cite{liu2019ppn,liu2019gpn}, an extra memory of representative feature maps~\cite{meta-mem-aug} and an optimization strategy for fast adaptation~\cite{ravi2017optimization}.
The ideas have been broadly used in applications for computer vision~\cite{schwartz2019repmet}, natural language processing~\cite{gu2018meta} and health~\cite{zhang19metapred} domains.

Most of the meta-learning based approaches for few-shot learning mainly target the task of image classification. Images can be collected from websites or camera devices and image datasets have vast amounts of data, e.g., ImageNet has more than 14 million hand-annotated images; The CIFAR-10 and CIFAR-100 are labeled subsets of the 80 million tiny images dataset~\cite{krizhevsky2009learning}.
However, time-series data has limited data sources and the annotations often require expertise knowledge, which has a higher cost. As a result, we define our problem as few-shot time-series learning on the small data area, where we focus on one task with limited samples instead of a continual learning setting with the idea of meta learning.
To the best of our knowledge, we are the first to focus on the problem of few-shot time-series learning.

\subsection{Interpretability for Neural-based Model}
Interpretability is critically important for neural-based models due to its black-box nature. Conventional model interpretation analyses the model’s parameters and outcomes to generate an interpretation for the model, for example,~\cite{fisher2018model}~analysis feature importance for generating results, explaining the model using prototype examples~\cite{kim2016examples} or influential examples \cite{koh2017understanding}, and discovering influential input patterns using activation maximisation criteria~\cite{montavon2018methods}. Gradient analysis is also important for model interpretation, e.g. Layer-Wise Relevance Propagation~\cite{bach2015pixel}, and gradient sensitivity analysis \cite{montavon2018methods}. Many research works have tried to modify the existing DNN/CNN/RNN-based or attention-based neural model to generate intermediate information for model interpretation. 

Another research direction for model interpretability is to treat the original model as a black-box, then use another interpretable model to simulate or approximate the original model without opening it. For example,~\cite{frosst2017distilling} distill a black box neural network into a soft decision tree that is an inherently interpretable model\cite{wu2018beyond} and generate a decision tree to interpret the RNN-based deep model for time series classification.

\section{Dual Prototypical Shapelet Networks} \label{sec:method}
Following most recent time series learning works~\cite{zhang2016unsupervised,schafer2017fast,bagnall2017great}, we learn from training dataset $\mathcal T =\{T_{1},...,T_{N}\}$ annotated by labels $\mathcal Y= \{y_{1},...,y_{N}\}$. $N$ is the number of training data. We use $K$ to denote the number of classes. We undertake classification in more challenging scenarios where the number of training samples for every class is reduced to 2, i.e., $\| \mathcal T^{k}\|$=2. We do not test when $\| \mathcal T^{k}\|$=1 for it makes no sense to learn a prototype from one sample without prior knowledge. 
The overall framework of the proposed DPSN method is shown in Figure ~\ref{fig:framework}.
The whole procedure is divided into two stages: feature extraction using SFA (Sec.~\ref{sec:sfa-feature}) and classification using a prototypical network (Sec.~\ref{sec:pp}). 

\begin{algorithm}
\caption{Training process for DPSN}
\label{alg:train-dpsn}
\textbf{Input:} SFA feature $\mathcal{X}$, Label $\mathcal{Y}$; Transformation net $f_{\phi}(\cdot)$. Number of samples for classification in every iteration $N_{q}$.

\textbf{Output:} Trained transformation net $f_{\phi}(\cdot)$; Prototype of each class $\{\mC_{k}\}_{k=1}^K$.
\begin{algorithmic}[1]
\STATE Initialize $f_{\phi}$ with random weight; $L\leftarrow0$
\WHILE{not converge}
\STATE Sample a batch of features $\mathcal{X}^{k}$ for every class $k \in \mathcal{Y}$
\STATE  Generate $\mC_{k}$ for every class using Equation~\ref{prototype_center}
\STATE  For each class $k$, randomly pick $N_{q}$ samples from $\mathcal{X}^{k}$
\STATE Calculate loss using these examples by Equation~\ref{eq:cross-entropy}
\STATE Apply Mini-batch SGD to minimize $L$ and update $\phi$,
\ENDWHILE

\end{algorithmic}
\end{algorithm}
\subsection{SFA Feature Extraction}
\label{sec:sfa-feature}
The symbolic Fourier approximation(SFA)~\cite{schafer2012sfa} word histogram is a handcrafted feature, which has achieved promising results in time series classification~\cite{schafer2017fast,schafer2015boss}. However, typically, the SFA word histogram is sparse and high-dimensional. Thus, we do some dimension reduction on the histogram. We remove those words from histogram if their values  are all 0 for each training data.
In testing, we do not count words, which have already been removed.
In brief, we transform the time series to dense SFA features i.e., $ \mathcal T \rightarrow \mathcal X =\{\vx_{1},...,\vx_{N}\} $, for classification below.

SFA word histogram requires two hyperparameter sampling window size $L_{S}$ and a $\vw$ for discrete Fourier transform (DFT) noise reduction.
We follow the method in~\cite{schafer2015boss} to pick these two hyperparameters.

\subsection{Prototypical Network}
\label{sec:pp}
We use a prototypical network, which is originally introduced to tackle the few-shot image classification task, to solve the few-shot time series classification problem. 
The prototype network is a KNN based classifier, which classifies samples of the class with the nearest prototype. 
Prototype $\mC_{k}$ for class $k$ is defined by the feature average of $\mathcal X^k$:
\begin{equation} \label{prototype_center}
\mC_{k}=\frac{1}{{\| \mathcal X^{k}\|}}\sum_{x\in \mathcal X^{k}}{f_{\phi }({{\vx})}}, 
\end{equation}
where $\mathcal X^k$ denotes SFA features of sampled training samples in class $k$, $\vx$ is SFA feature for a time series and $f(\cdot)$ is the feature extractor parameterised by $\phi$. 

Even though most time series networks use 1d CNN for feature transformation $f_{\phi}$, we prefer a fully connected layer to map the feature from the SFA domain into the prototype domain.
There are two reasons to use the SFA feature. Firstly, the SFA feature is a time invariance feature. Thus, we don't have to use CNN for it can obtain a time invariance feature. Secondly, the SFA feature is a handcrafted feature, which makes it is easy to know what frequencies are denoised during feature extraction. And this characteristic is essential for shapelet finding.

Classification result is calculated as a softmax over all classes $k'\in \mathcal{Y}$ using Euclidean distance between a sample and every prototypes:
\begin{equation} \label{distance_metric}
\Pr(y=k|\vx)=\frac{\exp (-\|{{f}_{\phi }}(\vx) - {{\mC}_{k}}\|^{2})}{\sum\nolimits_{{{k}^{'}}}{\exp (-\|{{f}_{\phi }}(\vx) - {{\mC}_{{{k}^{'}}}}\|^{2})}}
\end{equation}

The cross entropy loss $L$ of data sampled from the distribution $\mathcal{D}$ can be formulated as:

\begin{equation}\label{eq:cross-entropy}
    L={\mathbb E_{(\vx,y)\sim\mathcal D}-\log\Pr(y|\vx;\phi)}
\end{equation}
More details for this training strategy are shown in Algorithm.~\ref{alg:train-dpsn}.

\subsection{Interpret Model Using Representative Shapelets}
As discussed in the training process subsection, the classifier will discover an aggregated central point as a prototype that can mostly represent all samples in the class. Intuitively, if we could transform the SFA feature-based prototype to a time series format, we can obtain a good representative sample of the class. However, this operation is impossible because the SFA features are not derivative. Therefore, we choose a representative example by finding the time series example which is the nearest neighbor of the prototype in space $\phi$. This representative sample is the prototype approximation that can demonstrate the overall shapes of all supportive samples in the class. These representative shapelets for classes are a type of explanation for the trained DPSN classifier.

Given a class $k$, the prototype is generated by simply calculating the weighted sum of all supportive samples in the class, and the representative sample is selected using the following equation
\begin{equation}
     T_{i},  i = \arg \min_i F(f_{\phi}(x_{i}), \mC_{k})
\label{eq:disc-shape}
\end{equation}

where F is the distance function which could be Euclidean distance or another, e.g. cosine dissimilarity, or KL divergence. The distance calculation is based on the representation in space $\phi$. Thus, in our work, we use Euclidean distance as it is easy to understand by humans~\cite{zhang2018salient}.

\subsection{Interpret Model Using Discriminative Shapelets}
When people introduce or describe a particular object, they need to use specific or unique attributes that can enable the audience to distinguish the object from others. A time-series sample includes numerous sub-sequences, each of which consists of a shapelet and a noise. Given a class, its discriminative shapelet should 1) exist in every same-class sample, and 2) not exist in any diff-class sample. We denote the representative sample by $T$ and extract many sub-sequence $T^{i}$ using a slider window. The shapelet of $T^{i}$ is $S^{i}$, discriminative shapelets are denoted as $S$. 

\begin{algorithm} 
\caption{Discriminative Shapelet Discovery}

\textbf{Input:} Time series data$\mathcal T$, SFA feature $\mathcal X$, Label $\mathcal Y$, well trained linear transformation network $f_{\phi}(\cdot)$

\textbf{Output:} discriminative shapelet of each class $\{S_{k}\}_{i=1}^K$
\begin{algorithmic}[1]
\STATE calculating $\{C_{k}\}_{i=1}^K$ by Equation~\ref{prototype_center}
\FOR{$C_{k}$ in $\{C_{k}\}_{i=1}^K$} 
\STATE Identity $T$ by Equation~\ref{eq:disc-shape}
\FOR {each sub-sequence $T^{i}$ in $T$}
\STATE Applying DFT on $T^{i}$
\STATE Use the first $w$ complex number to reconstruct $S^{i}$
\STATE Calculate $D =[d_{S^{i}T_{1}},..., d_{S^{i}T_{N}}]$ by Equation~\ref{shapelet_distance} 
\STATE Calculate f-test score of $D$ and $Y^{k}$
\ENDFOR
\STATE save the $S^{i}$ with highest f-test score as $S_{k}$
\ENDFOR
\STATE \textbf{Return}
\end{algorithmic}
\label{alg:discriminative}
\end{algorithm}

\begin{table*}[ht!]
\centering
\resizebox{2\columnwidth}{!}{%
\begin{tabular}{|l|c|c|c|c|c|c|}
\hline
                                & \multicolumn{3}{c|}{6-shots Accuracy(std) \%}                      & \multicolumn{3}{c|}{8-shots Accuracy(std) \%}                      \\ \hline
Classifier(ways, original training size)                      & DPSN                 & BOSS                 & ST                   & DPSN                 & BOSS                 & ST                   \\ \hline
ArrowHead(3-way, 36)                & \textbf{78.725(3.9259)} & 67.6(7.5859)            & 51.2571(8.0994)          & \textbf{83.0306(2.1732)} & 74.4571(4.7889)          & 47.7714(10.0906)          \\ \hline
BME(3-way, 30)                      & 72.0(9.4157)          & \textbf{82.7333(2.1419)} & 64.0667(4.9933)          & 74.3583(8.9955)          & \textbf{86.1333(2.8769)} & 69.2(5.1621)            \\ \hline
CBF(3-way, 30)                      & \textbf{97.5778(1.759)} & 96.1(2.0684)          & 89.7556(1.7907)           & \textbf{97.225(1.8277)} & 95.9889(1.5418)          & 88.6333(2.2389)         \\ \hline
Chinatown(2-way, 20)                & \textbf{80.4083(3.2779)} & 78.5507(4.9493)           & 71.5942(19.356)        & \textbf{83.3792(2.2765)} & 81.6812(2.7862)          & 71.7101(17.5538)        \\ \hline
ECG200(2-way, 100)                   & \textbf{83.1667(4.2512)} & 75.9(5.6853)           & 65.6(3.1693)          & \textbf{86.5167(1.4359)} & 78.0(4.6904)           & 67.1(3.9285)           \\ \hline
GunPoint(2-way, 50)                 & \textbf{95.425(1.8069)} & 84.5333(4.6593)         & 79.8667(7.9181)         & \textbf{95.8542(1.8457)} & 88.0(3.4427)          & 58.7333(11.8819)        \\ \hline
GunPointAgeSpan(2-way, 135)          & \textbf{84.9167(3.5673)} & 80.5696(4.5056)         & 63.6392(15.8484)        & \textbf{90.5208(3.0109)} & 87.5316(4.0941)          & 83.7975(12.8906)        \\ \hline
GunPointOldVersusYoung(2-way, 135)   & \textbf{80.1417(5.0525)}  & 77.2381(6.2641)         & 68.8254(9.8393)         & \textbf{87.6208(3.908)}  & 85.3016(6.1486)          & 72.381(9.6546)         \\ \hline
ItalyPowerDemand(2-way, 67)         & \textbf{80.8917(6.5831)} & 78.8144(4.1512)          & 62.6142(12.0767)        & \textbf{84.9042(3.0184)} & 82.964(3.1474)        & 77.9397(6.1888)       \\ \hline
MoteStrain(2-way, 20)               & \textbf{82.4083(2.5158)} & 79.0256(2.2428)          & 77.8674(4.1452)          & \textbf{82.9417(2.6372)} & 80.615(1.2533)         & 79.7204(2.5927)        \\ \hline
Plane(7-way, 105)                    & \textbf{99.75(0.5586)} & 99.2381(0.7512)        & 55.4286(11.1861)         & 99.9345(0.2071)         & \textbf{99.5238(0.6734)} & 53.4286(9.9426)          \\ \hline
SonyAIBORobotSurface1(2-way, 20)    & 55.1208(2.7508)          & 53.7105(9.8365)         & \textbf{58.5025(7.2439)}  & 55.9292(2.9125)         & 54.193(9.8447)          & \textbf{58.8353(7.6576)} \\ \hline
SonyAIBORobotSurface2(2-way, 27)    & \textbf{81.2208(2.2143)}  & 79.7167(1.6636)         & 68.8982(2.9631)          & \textbf{84.0958(1.3932)}  & 83.064(1.8536)         & 75.0157(3.7848)          \\ \hline
SyntheticControl(6-way, 300)         & 76.4681(5.493)          & 76.5667(1.2673)         & \textbf{76.7333(9.0823)} & 83.9611(2.525)          & 79.8(1.7156)          & \textbf{87.5333(7.5559)} \\ \hline
ToeSegmentation1(2-way, 49)         & \textbf{96.1792(1.2086)} & 85.3947(3.6347)         & 77.5439(5.3309)         & \textbf{96.4792(0.8191)} & 85.9649(4.7823)          & 79.9123(4.3556)          \\ \hline
TwoLeadECG(2-way, 23)               & \textbf{93.4833(1.7817)} & 89.7981(3.1871)          & 75.8297(9.7762)       & \textbf{95.0125(0.5621)} & 91.4311(2.1031)          & 74.8376(5.1221)         \\ \hline
UMD(3-way, 36)                      & 85.6194(2.758)         & \textbf{85.9028(6.2804)}  & 85.5556(5.0376)          & 89.2167(4.3268)          & \textbf{90.9722(2.4715)} & 86.875(5.7026)          \\ \hline
Wine(2-way, 57)                     & \textbf{64.5167(7.1147)} & 57.2222(6.5006)          & 50.0(0.0)            & \textbf{66.3542(7.7271)} & 57.4074(4.0946)          & 50.0(0.0)            \\ \hline
average accuracy Rank           & \textbf{1.25}        & 2.25                 & 2.5                  & \textbf{1.19}        & 2.33                 & 2.48                 \\ \hline
average std Rank                & \textbf{1.55}        & 1.95                 & 2.5                  & \textbf{1.71}        & 1.86                 & 2.43                 \\ \hline
Wins                            & \textbf{14}          & 2                    & 4                    & \textbf{13}          & 3                    & 4                    \\ \hline
\end{tabular}
}
\caption{The accuracy(standard deviation) result of 18 datasets in 6-shot and 8-shot scenario, and statistical results of average accuracy and standard deviation rank. The wins row is the total number of best accuracy in 18 datasets}
\label{table:total}
\end{table*}

Given the class $k$, its best or optimal discriminative shapelet $S_{k}$ can be found by maximizing the likelihood in Equation~\ref{eq:likehood}.
\begin{equation}
\label{eq:likehood}
    S_{k} = \arg \max_{i} L_{o}(S^{i},\mathcal T,\mathcal Y^{k})
\end{equation}
where $\mathcal T$ and $\mathcal Y^{k}$ combine to product a bi-classification dataset  where the positive class is the given class $k$ and the negative class is all other classes.
The likelihood function needs to be aligned to the aforementioned two requirements of discriminative shapelets. In particular, we measure the likelihood using f-test as demonstrated in Equation \ref{eq:ftest}. 
\begin{equation}
L_{o} =\frac{between\ class\ variability}{within\ class\ variability}
\label{eq:ftest}
\end{equation}
where the large value of ``between-class variability" indicates that the selected shapelet does ``not exist" in the diff-class samples, and the small value of the ``within-class variability" indicates that the shapelet is likely to ``exist" in the same-class samples. The shapelet has the largest f-test value is the discriminative shapelet.

The ``between-class variability" measures the distance between different classes using the following equation:
\begin{equation}
    \frac{\sum_{i=1}^K N_i*(\overline{d}_i - \overline{d})^2}{K-1}
\end{equation}
where $N_i$ is the number of samples i the $i-th$ class, $\overline{d}_i$ denotes the mean value of all samples in the $i-th$ class, $\overline{d}$ denotes the overall mean of all classes, and $K$ is the number of classes that is 2 in the bi-classification scenario.

The ``within-class variability" calculates the variance of all the distance-values in the same class using the following equation.
\begin{equation}
    \frac{\sum_{i=1}^K \sum_{j=1}^{N_i} (d_{ij} - \overline{d}_i)^2}{N-K}
\end{equation}

As discussed, $d$ is the value to estimate the likelihood. $d$ is the distance between a shapelet and a time-series example. In particular, a time-series $T$ will be transformed to a set of subsequences $T^{i}$ that is the same length as shapelet $S$. As we expect to check whether or not the shapelet exists in the time-series, distance $d_{ST}$ is the minimal distance between $S$ and the set of subsequences $T^{i}$.
as calculated by Equation \ref{shapelet_distance}.
\begin{equation}\label{shapelet_distance}
d_{ST} =\min_{i=1\dots ns} F(T^{i}, S)
\end{equation}
where $F$ is the distance function, $ns$ is the number of generated sub-sequences from time-series $T$.

\subsubsection{Procedure for discovering discriminative shapelets}
To discover the discriminative shapelets, we use the following four steps.
\begin{itemize}
    \item Extract sub-sequences from the selected time series data using a sliding window.
    \item Extract signals from the sub-sequences using DFT. 
    \item Measure the shapelet distance between each signal and the time series in the training data.
    \item Use f-test to find the best signal as the discriminative shapelets.
\end{itemize}

The pseudo code for discovering discriminative shapelets is described in Algorithm \ref{alg:discriminative}

\begin{figure*}[ht]
\centering
\includegraphics[width=0.75\linewidth]{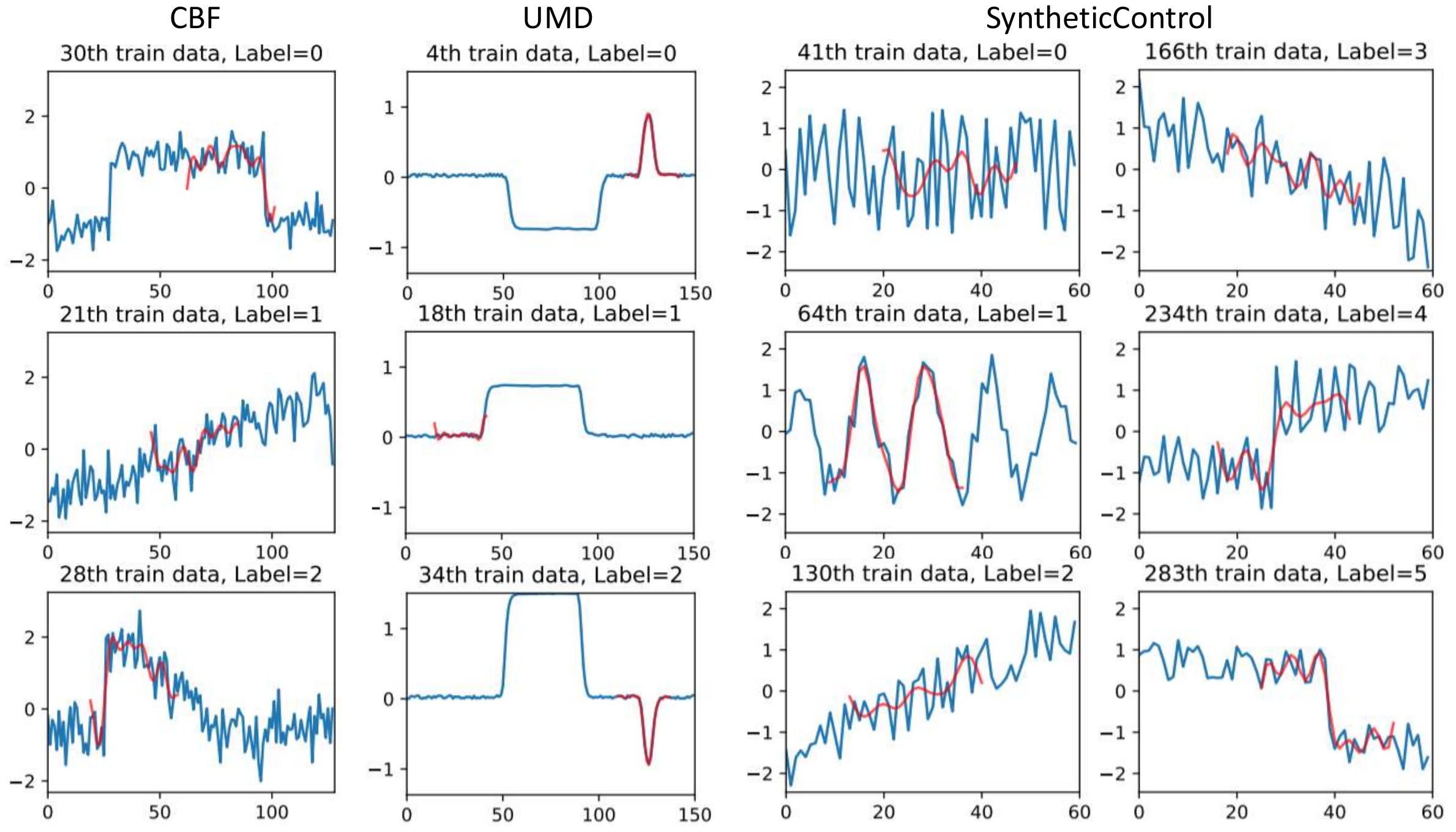}
\caption{\label{fig:interResult} 
This figure shows the interpretability results of three handcrafted dataset. Names of the datasets are above each graph. On the top of each line chart, the title has two numbers $i$th and label=$j$. These numbers indicate that the $i$th data in the training set is selected to represent class $j$. The blue line is the $i$th data, while the red line is the discriminative shapelet from our framework.
}
\end{figure*}

\begin{figure}[ht]
    \centering
    \includegraphics[width=0.75\linewidth]{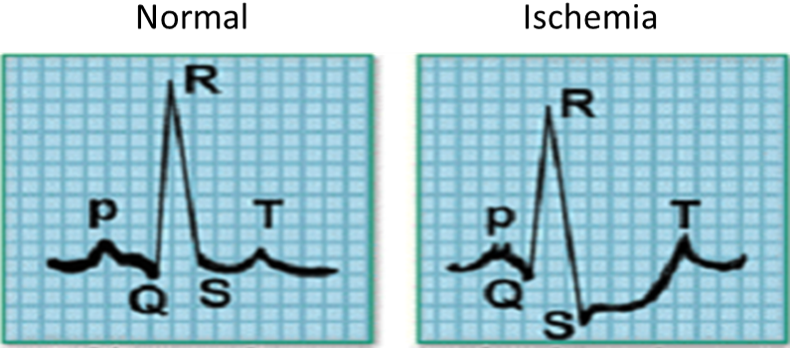}
    \caption{The visualization of differences between Normal and Ischemia on ECG200~\cite{olszewski2001generalized} from UCR archive~\cite{UCRArchive2018}. The lower value of S and long time from S to T is called ST-segment depression, and lower value of R is called invalid amplitude}
    \label{fig:ecg_normal_vs_ischemia}
\end{figure}

\begin{figure}[ht]
    \centering
    \includegraphics[width=0.75\linewidth]{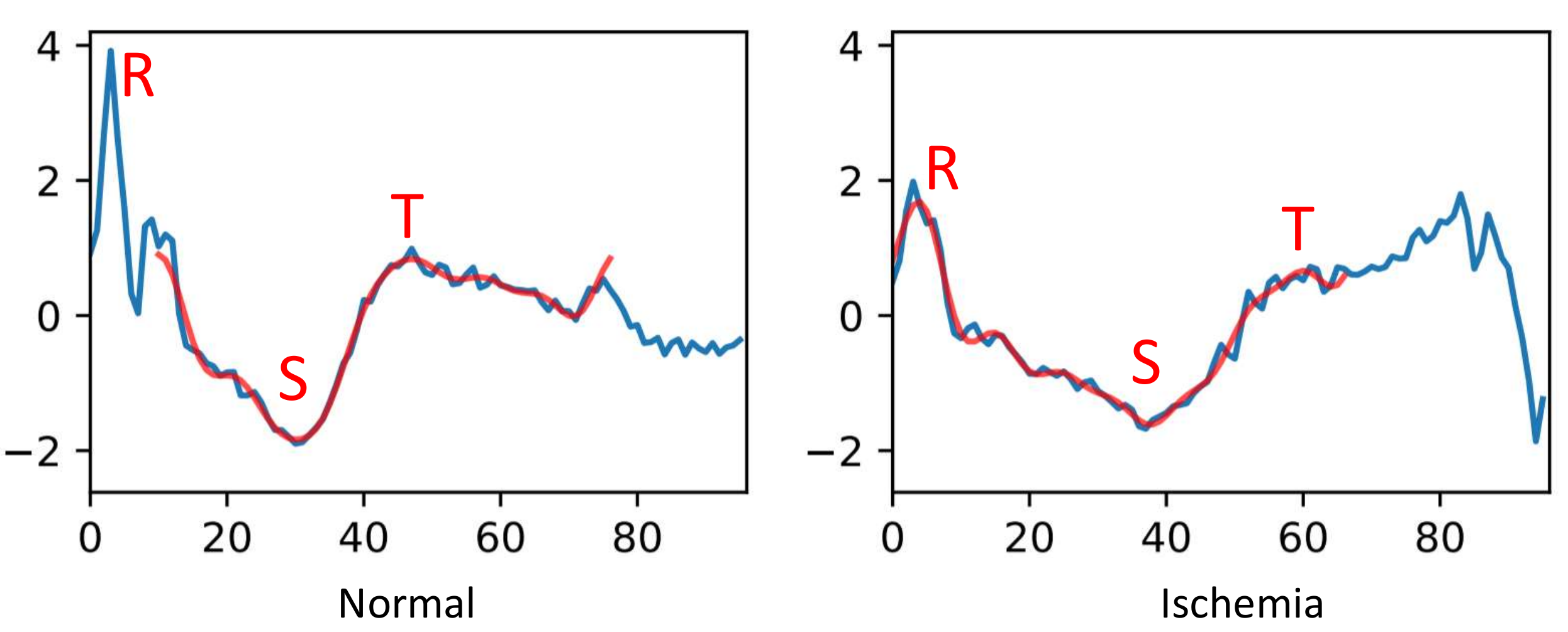}
    \caption{The visualization of differences between Normal and Ischemia, from dataset description of ECG200 datasets [40], by UCR archive[12], 2015, URL:\url{http://www.timeseriesclassification.com/description.php?Dataset=ECG200}. Copyright 2015 by UCR archive}
    \label{fig:ecg_shapelets}
\end{figure}

\section{Experiments} \label{sec:experiments}
\subsection{Dataset Selection}
To experimentally study the proposed model, we chose and derived 18 few-shot time-series classification datasets from UCR \cite{UCRArchive2018} dataset which included 128 time-series datasets. In particular, we prefer dataset with limited training samples and short length for each time series. We rank the 128 UCR datasets from UCR by the $length \times size\ of\ the\ training\ sample$ and choose the top 18 datasets to be converted to few-shot sample scenario. 

The sampling procedure needs a hyper-parameter named sample ratio which is expressed as a fraction of the total number of training samples. Moreover, for each sample ratio, we will sample them 10 times.
In this experiment, we set sample ratio number from 0.1 to 0.9 with the step of 0.1. Thus, for one UCR dataset, we can build $9\times 10$ few-shot datasets from it. We experimented on each few-shot datasets and also on the original UCR dataset. 

Specifically, assume the sample ratio is 0.3, and the original dataset has 100 training data, we will randomly sample 30 training data ten times. We will have 10 new datasets and each dataset has 30 training data.
The test datasets were kept the same as the test dataset in UCR benchmark.

We do not pre-process the input data, except to replace NaN value with zero for any unequal time series data.

\subsection{Baselines}
Two closely related baselines are used in our experiments.
We compare our method with BOSS\cite{schafer2015boss}, which also uses SFA histograms.
We also compare our method to Shapelet-Transform (ST)\cite{hills2014classification}, which also uses shapelets distance to extract the discriminative feature. The source code for BOSS and SFA feature extraction is seen in \footnote{\url{https://github.com/johannfaouzi/pyts}}. The source code of ST from 
\footnote{\url{https://github.com/rtavenar/tslearn/blob/master/tslearn/docs/index.rst}}.
\subsection{Experiment Setup}
Both SFA feature and ST need hyper-parameters like window size and etc. In the original code from UCR dataset, they use cross validation to select the best hyper-parameters. However, for few-shot datasets, the number of training samples can be very small. Thus, it is impossible to do cross validation to get those hyper-parameter. Thus, we use the same hyper-parameter as the cross validation selected in the total dataset.

For prototypical neural networks, the transformation network has two fully connected layers with the neuron number of hidden layer as 256 and the output dimension as 64.
We use Adam to train our model for 1000 epochs with the learning rate of 0.002, the learning rate decay of 1e-5, and the momentum of 0.7.
We published our code here\footnote{\url{https://github.com/Wensi-Tang/DPSN}}.

\subsection{Experiment Results}\label{sec:Experiment Result}

\subsubsection{Experiment Result of Classification}\label{sec:no_need_to_prototype}
In Tab~\ref{table:total}, we demonstrate the 6-shots and 8 shots result of selected 18 few-shot time-series datasets. We can see that when less training data is used, for more than half of 18 datasets, DPSN outperforms baselines on accuracy with lower variance. All results can be found on this link 
\footnote{\url{https://github.com/Wensi-Tang/DPSN/blob/master/results/result.txt}}.

\subsubsection{Interpretability of DPSN}
For all figures in this paper, horizontal and vertical axes are the same as UCR datasets, and the unit is not labeled as it is not important for the classification task~\cite{UCRArchive2018}.
We demonstrate the interpretability of DPSN in Figure~\ref{fig:interResult}. We choose three simulated datasets. Those datasets are chosen because they are easy for humans to understand, and in the following section, we will demonstrate the interpretability of DPNS by using two real case studies. A visualization of all results can be found on this link 
\footnote{\url{https://github.com/Wensi-Tang/DPSN/tree/master/SFA_Python-master/test/image_result}}.

As it is shown in Fig~\ref{fig:interResult}, the shapelet we find is not a subsequence from the original datasets. The discriminative shapelets is a de-noised shapelets from the original shapelets. Our method is based on the SFA feature and SFA feature drops the information on high frequencies. Therefore, when we find the discriminative shapelets, we also drop that high frequency information that the feature cannot see. Therefore, the shapelets we demonstrate is what the feature sees and what the model uses to achieve classification. 

\begin{figure*}
\centering
\includegraphics[width=0.75\linewidth]{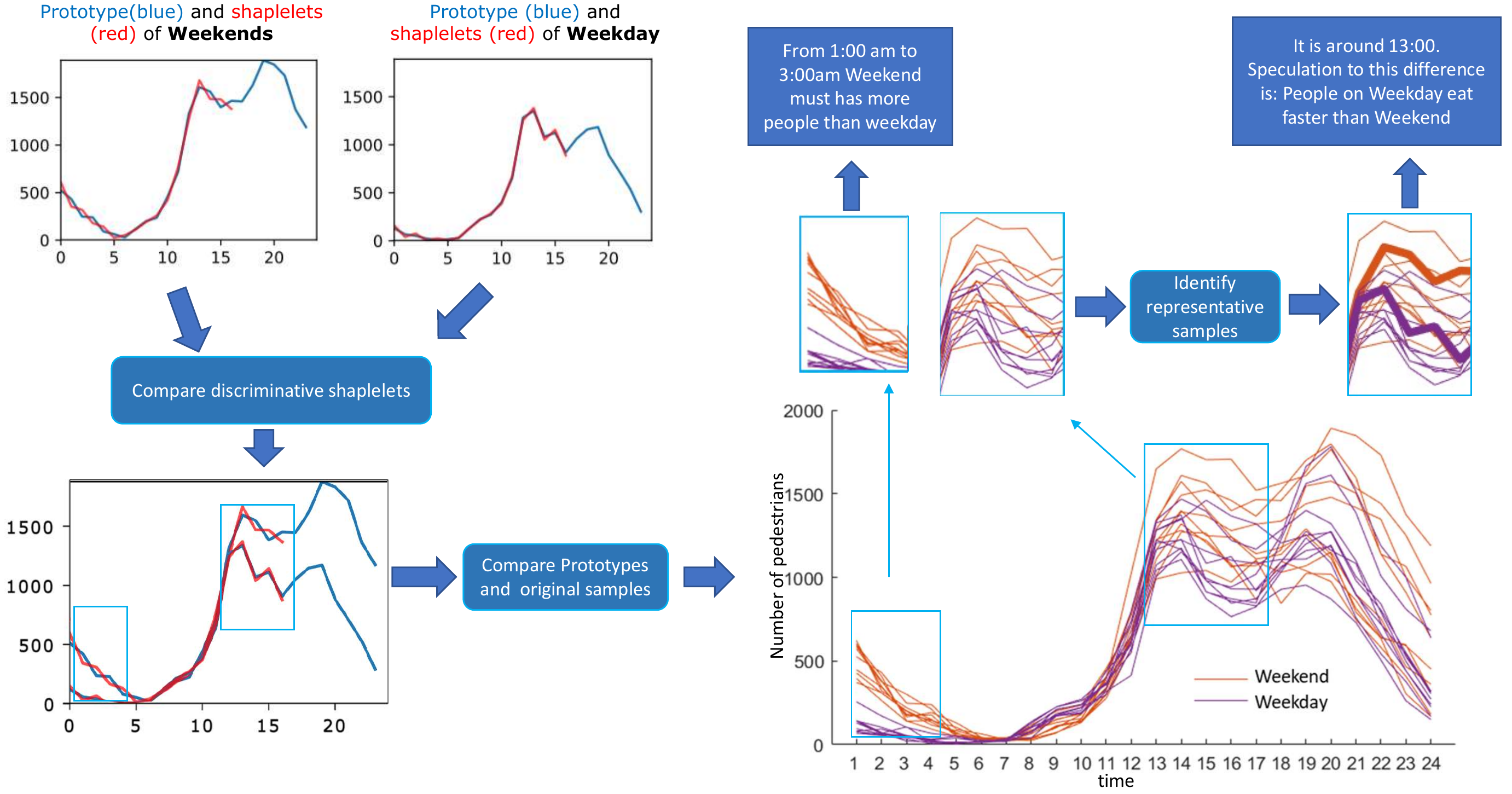}
\caption[Caption for LOF]{A comparison of shapelets and original samples for Chinatown dataset. The left side shows the prototypes for pedestrians on weekends or weekdays. The important features of the prototypes, which are indicated by the boxes are carefully analyzed together with the original samples on the right side. If you are interested in the confirmation of the speculation, here is a \href{https://github.com/Wensi-Tang/DPSN/blob/master/Appendix/chinatwon_speculation_and_confirmation.pdf}{\textcolor{blue}{\underline{link}}}.}\label{fig:china-town-inter}
\end{figure*}

\subsection{Classification Result Analysis}\label{sec:Classification_Result_Analysis}

The UCR branch has time-series datasets from various sources; thus, it is hard to develop one classification method which could outperform every technique in every dataset. Therefore, in this part, we want to analyze the relationship between DPSN's performance and the characteristic of time-series. This analysis will benefit model selection.

From a visualization of full result on this link\footnote{\url{https://github.com/Wensi-Tang/DPSN/blob/master/All_classification_result.pdf}}, we could see that DPSN has a great relationship with BOSS. This relationship can also be seen in Tab.~\ref{table:total}. The reason is that despite redesigning the distance metric method and classifier, the SFA feature histogram which we use is the same as BOSS. This means that will inherit problems from the SFA feature. 

Furthermore, the problem is that if a dataset can be described by the SFA feature, which characteristics of a dataset would help us make a selection between BOSS or DPSN. When compared with the 6-shot and 8-shot results, we could see that only for the BME dataset, the BOSS is always superior to DPSN. This is because the BME dataset has more than one cluster in its class. Thus, learning only one prototype for each class is not sufficient. This kind of problem can be solved by the method described in~\cite{allen2019infinite} 

Another phenomenon is that despite the expectations of the result of 8-shot to be better than the results of 6-shot, it not true for all datasets. For example, as Tab~\ref{table:total} shows, for both accuracy and standard deviation, the 6-shot results of Plan dataset is better than that of the 8-shot. This phenomenon can also be found in other works of few-shot learning tasks like the one in\cite{huiwenshi}. The reason is that the relationship between performance and size of training datasets is not a monotonic function. There are some fluctuations in it.

From the visualization of all results we can find that the DPSN outperforms BOSS and ST for all sample ratios. Which means that the DPSN can be viewed as a general method. However, the reason we still claim our method, is a few-shot method is that the technique is a type of K-NN classifier, which practically is not an ideal solution for a dataset consisting of a large amount of training data.

\section{Case studies}
This section elaborates the interpretability of our model using several case studies in real-world datasets and applications.
\subsection{Heartbeat Anomaly Interpretation}
In the ECG200 time-series dataset \cite{olszewski2001generalized}, a binary classifier can be trained to detect the anomalous heartbeat time-series shapelets. As shown in Fig.~\ref{fig:ecg_normal_vs_ischemia}, there are two demonstrations for different shapelets of two kinds of heartbeat shapelets: normal heartbeat and Myocardial Infarction (Ischemia). We can find two shapelets that share a similar pattern which starts from a low value $Q$, then goes to a sharp peak $P$ and returns back to a low value $S$. Differences between normal and Ischemia lies on that Ischemia has a longer time from $S$ to $T$ (another high point after $S$), and a lower value of $R$ than a normal heartbeat, which is called invalid amplitude.

Our DPSN model produces one prototype for each class. As shown in  Fig.~\ref{fig:ecg_shapelets}, the left prototype is a normal heartbeat signal that can quickly recover from S to T while the Ischemia signal has a relatively slow recovery curve in the heartbeat shapelets. Moreover, the discriminative shapelets found by DPSN, which is highlighted by red lines match the sub-sequence between R and T in the demonstrated example in Fig.~\ref{fig:ecg_normal_vs_ischemia}. 
The recovery time from $S$ to $T$ is prolonged in the case of Ischemia on the right side. The value of $R$ for Ischemia is lower than the normal heartbeat case.

We can see that the representative shapelets of ischemia is smooth when compared with that of normal class. This phenomenon can also be found in the class 0 in Fig~\ref{fig:ecg_normal_vs_ischemia}. The reason is that DPSN is based on the SFA feature. To be specific, the SFA feature drops the information on high frequencies, and when we finds the representative shapelets, we also drop those high frequency information that the SFA feature cannot see. The shapelets we demonstrate is what the feature see. Therefore, the smooth shapelet happens just because those representative samples are of clean frequency characteristic.

\subsection{Analysis of Pedestrian Patterns on Weekday and Weekend}

The City of Melbourne, Australia has deployed a pedestrian counting system to automatically count pedestrian flow within the city and to gain deeper understanding of public resource usage in multiple city locations at different times of the day. Data analysis can facilitate the decision maker to make better urban planning decisions for future city development in a human-centric way. As shown in the original features of the dataset, we can draw some interesting conclusions: 1) in an urban area, e.g. Chinatown, there are more pedestrians during the weekdays because many people work in urban areas and stay in theri home suburbs on the weekends. 2) pedestrians are more likely to have a short lunch-time during weekdays and spend more time on the lunch during weekends.

As demonstrated in Figure~\ref{fig:china-town-inter}, the prototypes for each class generated from our model shows the above patterns. Moreover, two partial-length shapelets are extracted to identify the most discriminative characteristics. We found that there are many pedestrians at midnight on the weekends whilst there are hardly any people on the streets at midnight during the weekdays. This characteristic of overnight activities can be used as important features to classify the weekends and weekdays. The characteristics of populations, as mentioned above, are not the best discriminative feature. Because the population in a location is usually impacted by events that will eventually bring noise to this feature. In summary, the prototypes can demonstrate the most obvious patterns of the data, however, the discriminative shapelets can reveal the most important feature for classification.

\section{Conclusions}\label{sec:Conclusions}
We propose the novel Dual Prototypical Shapelet Networks model for interpretable time-series classification on few-shot samples. In experiments, DPSN outperforms handcraft distance measurement, and the proposed model can be interpreted on dual granularity: representative sample with overview and discriminative shapelets with detail view. We use two case studies to demonstrate how the interpretability can facilitate the model explanation and derive new knowledge for data analysis. 

\bibliographystyle{ieeetr}
\bibliography{ref}
\end{document}

%% file: math_commands.tex
\usepackage{amsmath,amsfonts,bm}

\def\eqref#1{equation~\ref{#1}}

\def\1{\bm{1}}

\def\vw{{\bm{w}}}
\def\vx{{\bm{x}}}

\def\mC{{\bm{C}}}

\DeclareMathAlphabet{\mathsfit}{\encodingdefault}{\sfdefault}{m}{sl}
\SetMathAlphabet{\mathsfit}{bold}{\encodingdefault}{\sfdefault}{bx}{n}













%% file: main.bbl
\begin{thebibliography}{10}

\bibitem{busatto2008voxel}
G.~F. Busatto, B.~S. Diniz, and M.~V. Zanetti, ``Voxel-based morphometry in
  alzheimer’s disease,'' {\em Expert review of neurotherapeutics}, vol.~8,
  no.~11, pp.~1691--1702, 2008.

\bibitem{sun2017effects}
J.~C.-Y. Sun and K.~P.-C. Yeh, ``The effects of attention monitoring with eeg
  biofeedback on university students' attention and self-efficacy: The case of
  anti-phishing instructional materials,'' {\em Computers \& Education},
  vol.~106, pp.~73--82, 2017.

\bibitem{chen2018a}
W.-Y. Chen, Y.-C. Liu, Z.~Kira, Y.-C.~F. Wang, and J.-B. Huang, ``A closer look
  at few-shot classification,'' in {\em ICLR}, 2019.

\bibitem{snell2017prototypical}
J.~Snell, K.~Swersky, and R.~Zemel, ``Prototypical networks for few-shot
  learning,'' in {\em Advances in Neural Information Processing Systems},
  pp.~4077--4087, 2017.

\bibitem{NIPS2017_7244}
S.~Motiian, Q.~Jones, S.~Iranmanesh, and G.~Doretto, ``Few-shot adversarial
  domain adaptation,'' in {\em Advances in Neural Information Processing
  Systems 30} (I.~Guyon, U.~V. Luxburg, S.~Bengio, H.~Wallach, R.~Fergus,
  S.~Vishwanathan, and R.~Garnett, eds.), pp.~6670--6680, Curran Associates,
  Inc., 2017.

\bibitem{long2012tcsst}
G.~Long, L.~Chen, X.~Zhu, and C.~Zhang, ``Tcsst: transfer classification of
  short \& sparse text using external data,'' in {\em Proceedings of the 21st
  ACM international conference on Information and knowledge management},
  pp.~764--772, 2012.

\bibitem{jiang2020decentralized}
J.~Jiang, S.~Ji, and G.~Long, ``Decentralized knowledge acquisition for mobile
  internet applications,'' {\em World Wide Web}, pp.~1--17, 2020.

\bibitem{wahbeh2007binaural}
H.~Wahbeh, C.~Calabrese, H.~Zwickey, and D.~Zajdel, ``Binaural beat technology
  in humans: a pilot study to assess neuropsychologic, physiologic, and
  electroencephalographic effects,'' {\em The Journal of Alternative and
  Complementary Medicine}, vol.~13, no.~2, pp.~199--206, 2007.

\bibitem{gao2014analysis}
X.~Gao, H.~Cao, D.~Ming, H.~Qi, X.~Wang, X.~Wang, R.~Chen, and P.~Zhou,
  ``Analysis of eeg activity in response to binaural beats with different
  frequencies,'' {\em International Journal of Psychophysiology}, vol.~94,
  no.~3, pp.~399--406, 2014.

\bibitem{ratanamahatana2005three}
C.~A. Ratanamahatana and E.~Keogh, ``Three myths about dynamic time warping
  data mining,'' in {\em Proceedings of the 2005 SIAM international conference
  on data mining}, pp.~506--510, SIAM, 2005.

\bibitem{schafer2015boss}
P.~Sch{\"a}fer, ``The boss is concerned with time series classification in the
  presence of noise,'' {\em Data Mining and Knowledge Discovery}, vol.~29,
  no.~6, pp.~1505--1530, 2015.

\bibitem{UCRArchive2018}
H.~A. Dau, E.~Keogh, K.~Kamgar, C.-C.~M. Yeh, Y.~Zhu, S.~Gharghabi, C.~A.
  Ratanamahatana, Yanping, B.~Hu, N.~Begum, A.~Bagnall, A.~Mueen, and
  G.~Batista, ``The ucr time series classification archive,'' October 2018.
\newblock \url{https://www.cs.ucr.edu/~eamonn/time_series_data_2018/}.

\bibitem{fawaz2019deep}
H.~I. Fawaz, G.~Forestier, J.~Weber, L.~Idoumghar, and P.-A. Muller, ``Deep
  learning for time series classification: a review,'' {\em Data Mining and
  Knowledge Discovery}, pp.~1--47, 2019.

\bibitem{tang2020rethinking}
W.~Tang, G.~Long, L.~Liu, T.~Zhou, J.~Jiang, and M.~Blumenstein, ``Rethinking
  1d-cnn for time series classification: A stronger baseline,'' {\em arXiv
  preprint arXiv:2002.10061}, 2020.

\bibitem{fei2006one}
L.~Fei-Fei, R.~Fergus, and P.~Perona, ``One-shot learning of object
  categories,'' {\em IEEE Transactions on Pattern Analysis and Machine
  Intelligence (T-PAMI)}, 2006.

\bibitem{lampert2013attribute}
C.~H. Lampert, H.~Nickisch, and S.~Harmeling, ``Attribute-based classification
  for zero-shot visual object categorization,'' {\em IEEE transactions on
  pattern analysis and machine intelligence}, vol.~36, no.~3, pp.~453--465,
  2013.

\bibitem{imagenet}
J.~Deng, W.~Dong, R.~Socher, L.-J. Li, K.~Li, and L.~Fei-Fei, ``{ImageNet: A
  Large-Scale Hierarchical Image Database},'' in {\em IEEE Conference on
  Computer Vision and Pattern Recognition (CVPR)}, 2009.

\bibitem{He_2016_CVPR}
K.~He, X.~Zhang, S.~Ren, and J.~Sun, ``Deep residual learning for image
  recognition,'' in {\em IEEE Conference on Computer Vision and Pattern
  Recognition (CVPR)}, 2016.

\bibitem{finn2017model}
C.~Finn, P.~Abbeel, and S.~Levine, ``Model-agnostic meta-learning for fast
  adaptation of deep networks,'' in {\em Proceedings of the 34th International
  Conference on Machine Learning-Volume 70}, pp.~1126--1135, JMLR. org, 2017.

\bibitem{liu2019ppn}
L.~Liu, T.~Zhou, G.~Long, J.~Jiang, L.~Yao, and C.~Zhang, ``Prototype
  propagation networks ({PPN}) for weakly-supervised few-shot learning on
  category graph,'' in {\em International Joint Conferences on Artificial
  Intelligence (IJCAI)}, 2019.

\bibitem{liu2019gpn}
L.~Liu, T.~Zhou, G.~Long, J.~Jiang, and C.~Zhang, ``Learning to propagate for
  graph meta-learning,'' in {\em Neural Information Processing Systems
  (NeurIPS)}, 2019.

\bibitem{meta-mem-aug}
A.~Santoro, S.~Bartunov, M.~Botvinick, D.~Wierstra, and T.~Lillicrap,
  ``Meta-learning with memory-augmented neural networks,'' in {\em
  International Conference on Machine Learning (ICML)}, 2016.

\bibitem{ravi2017optimization}
S.~Ravi and H.~Larochelle, ``Optimization as a model for few-shot learning,''
  in {\em International Conference on Learning Representations (ICLR)}, 2017.

\bibitem{schwartz2019repmet}
E.~Schwartz, L.~Karlinsky, J.~Shtok, S.~Harary, M.~Marder, S.~Pankanti,
  R.~Feris, A.~Kumar, R.~Giries, and A.~M. Bronstein, ``Repmet:
  Representative-based metric learning for classification and one-shot object
  detection,'' in {\em The IEEE Conference on Computer Vision and Pattern
  Recognition (CVPR)}, 2019.

\bibitem{gu2018meta}
J.~Gu, Y.~Wang, Y.~Chen, K.~Cho, and V.~O. Li, ``Meta-learning for low-resource
  neural machine translation,'' in {\em Conference on Empirical Methods in
  Natural Language Processing (EMNLP)}, 2018.

\bibitem{zhang19metapred}
S.~X. Zhang, F.~Tang, H.~Dodge, J.~Zhou, and F.~Wang, ``Metapred: Meta-learning
  for clinical risk prediction with limited patient electronic health
  records,'' in {\em ACM SIGKDD Conference ON Knowledge Discover and Data
  Mining (KDD)}, 2019.

\bibitem{krizhevsky2009learning}
A.~Krizhevsky, ``Learning multiple layers of features from tiny images,'' tech.
  rep., Citeseer, 2009.

\bibitem{fisher2018model}
A.~Fisher, C.~Rudin, and F.~Dominici, ``Model class reliance: Variable
  importance measures for any machine learning model class, from the”
  rashomon” perspective,'' {\em arXiv preprint arXiv:1801.01489}, 2018.

\bibitem{kim2016examples}
B.~Kim, R.~Khanna, and O.~O. Koyejo, ``Examples are not enough, learn to
  criticize! criticism for interpretability,'' in {\em Advances in Neural
  Information Processing Systems}, pp.~2280--2288, 2016.

\bibitem{koh2017understanding}
P.~W. Koh and P.~Liang, ``Understanding black-box predictions via influence
  functions,'' in {\em Proceedings of the 34th International Conference on
  Machine Learning-Volume 70}, pp.~1885--1894, JMLR. org, 2017.

\bibitem{montavon2018methods}
G.~Montavon, W.~Samek, and K.-R. M{\"u}ller, ``Methods for interpreting and
  understanding deep neural networks,'' {\em Digital Signal Processing},
  vol.~73, pp.~1--15, 2018.

\bibitem{bach2015pixel}
S.~Bach, A.~Binder, G.~Montavon, F.~Klauschen, K.-R. M{\"u}ller, and W.~Samek,
  ``On pixel-wise explanations for non-linear classifier decisions by
  layer-wise relevance propagation,'' {\em PloS one}, vol.~10, no.~7,
  p.~e0130140, 2015.

\bibitem{frosst2017distilling}
N.~Frosst and G.~Hinton, ``Distilling a neural network into a soft decision
  tree,'' {\em arXiv preprint arXiv:1711.09784}, 2017.

\bibitem{wu2018beyond}
M.~Wu, M.~C. Hughes, S.~Parbhoo, M.~Zazzi, V.~Roth, and F.~Doshi-Velez,
  ``Beyond sparsity: Tree regularization of deep models for interpretability,''
  in {\em Thirty-Second AAAI Conference on Artificial Intelligence}, 2018.

\bibitem{zhang2016unsupervised}
Q.~Zhang, J.~Wu, H.~Yang, Y.~Tian, and C.~Zhang, ``Unsupervised feature
  learning from time series.,'' in {\em IJCAI}, pp.~2322--2328, 2016.

\bibitem{schafer2017fast}
P.~Sch{\"a}fer and U.~Leser, ``Fast and accurate time series classification
  with weasel,'' in {\em Proceedings of the 2017 ACM on Conference on
  Information and Knowledge Management}, pp.~637--646, ACM, 2017.

\bibitem{bagnall2017great}
A.~Bagnall, J.~Lines, A.~Bostrom, J.~Large, and E.~Keogh, ``The great time
  series classification bake off: a review and experimental evaluation of
  recent algorithmic advances,'' {\em Data Mining and Knowledge Discovery},
  vol.~31, no.~3, pp.~606--660, 2017.

\bibitem{schafer2012sfa}
P.~Sch{\"a}fer and M.~H{\"o}gqvist, ``Sfa: a symbolic fourier approximation and
  index for similarity search in high dimensional datasets,'' in {\em
  Proceedings of the 15th International Conference on Extending Database
  Technology}, pp.~516--527, ACM, 2012.

\bibitem{zhang2018salient}
Q.~Zhang, J.~Wu, P.~Zhang, G.~Long, and C.~Zhang, ``Salient subsequence
  learning for time series clustering,'' {\em IEEE transactions on pattern
  analysis and machine intelligence}, vol.~41, no.~9, pp.~2193--2207, 2018.

\bibitem{olszewski2001generalized}
R.~T. Olszewski, ``Generalized feature extraction for structural pattern
  recognition in time-series data,'' tech. rep., CARNEGIE-MELLON UNIV
  PITTSBURGH PA SCHOOL OF COMPUTER SCIENCE, 2001.

\bibitem{hills2014classification}
J.~Hills, J.~Lines, E.~Baranauskas, J.~Mapp, and A.~Bagnall, ``Classification
  of time series by shapelet transformation,'' {\em Data Mining and Knowledge
  Discovery}, vol.~28, no.~4, pp.~851--881, 2014.

\bibitem{allen2019infinite}
K.~R. Allen, E.~Shelhamer, H.~Shin, and J.~B. Tenenbaum, ``Infinite mixture
  prototypes for few-shot learning,'' {\em arXiv preprint arXiv:1902.04552},
  2019.

\bibitem{huiwenshi}
E.~Triantafillou, T.~Zhu, V.~Dumoulin, P.~Lamblin, U.~Evci, K.~Xu, R.~Goroshin,
  C.~Gelada, K.~Swersky, P.-A. Manzagol, {\em et~al.}, ``Meta-dataset: A
  dataset of datasets for learning to learn from few examples,'' {\em arXiv
  preprint arXiv:1903.03096}, 2019.

\end{thebibliography}
